\documentclass{article}
\usepackage{spconf,amsmath,graphicx}
\usepackage{times}
\usepackage{epsfig}
\usepackage{graphicx}
\usepackage{amsmath}
\usepackage{amssymb}
\usepackage{amssymb}
\usepackage{amsmath,amssymb,amsfonts}
\usepackage{algorithmic}
\usepackage{booktabs}
\usepackage{textcomp}
\usepackage{xcolor}
\usepackage{subcaption}
\usepackage{mathtools}
\usepackage{datetime}
\usepackage[margin=1.0in]{geometry}
\usepackage{xcolor}
\usepackage{longtable}
\usepackage{comment}
\usepackage{adjustbox}
\usepackage{enumitem}
\usepackage{amsmath}
\DeclareMathOperator{\E}{\mathbb{E}}
\usepackage{hyperref}
\usepackage[numbers]{natbib}
\setlist[itemize]{noitemsep, topsep=0pt}
\usepackage{enumitem}

\title{Generating Thermal Human Faces for Physiological Assessment using Thermal Sensor Auxiliary Labels}
\name{Catherine Ordun$^{\dagger}$ \qquad Edward Raff $^{\star,\dagger}$ \qquad Sanjay Purushotham$^{\dagger}$}
  
\address{$^{\dagger}$University of Maryland, Baltimore County\\$^{\star}$Booz Allen Hamilton}

\begin{document}

%
\maketitle

\begin{abstract}
Thermal images reveal medically important physiological information about human stress, signs of inflammation, and emotional mood that cannot be seen on visible images. Providing a method to generate thermal faces from visible images would be highly valuable for the telemedicine community in order to show this medical information. To the best of our knowledge, there are limited works on visible-to-thermal (VT) face translation, and
many current works go the opposite direction to generate visible faces from thermal surveillance images (TV) for law enforcement applications. As a result, we introduce \textbf{favtGAN}, a VT GAN which uses the pix2pix image translation model with an auxiliary sensor label prediction network for generating thermal faces from visible images. Since most TV methods are trained on only one data source drawn from one thermal sensor, we combine datasets from faces and cityscapes. These combined data are captured from similar sensors in order to bootstrap the training and transfer learning task, especially valuable because visible-thermal face datasets are limited. Experiments on these combined datasets show that favtGAN demonstrates an increase in SSIM and PSNR scores of generated thermal faces, compared to training on a single face dataset alone.
\end{abstract}

\section{Introduction}
\label{sec:intro}

Thermal imaging can support medical AI as a diagnostic tool since decades of physiological research have shown that temperatures in facial thermograms are highly correlated to vital measures and reveal signs of stress and inflammation, otherwise hidden on visible images \cite{ioannou2014thermal,selinger2006appearance,buddharaju2007physiology,pavlidis2000imaging,wilder1996comparison, goulart2019emotion, puri2005stresscam}. However, due to time-consuming data collection for human subjects using thermal cameras, there is usually insufficient thermal medical data to train any AI system for thermal imaging. 
To the best of our knowledge, we introduce the first $\mathit{visible} \rightarrow \mathit{thermal}$ (VT) facial translation method, which can be used to bootstrap or augment thermal training data. 
The ability to generate a thermal face image from a visible face may be challenging due to the fact that there are limited, paired thermal-visible face datasets \cite{ordun2020use}. Therefore, the ability to bootstrap the image-to-image translation task with additional training data from similar thermal sensors that share the same optical properties would improve training, even if the data are from different domains (e.g. faces and cityscapes).  We train on combined visible-thermal datasets in our method, called \textbf{favtGAN} (facial-visible-thermal-GAN  or ``favorite GAN"). favtGAN is a simple Generative Adversarial Network (GAN) architecture built on the popular pix2pix framework ~\cite{isola2017image}. Unlike pix2pix, favtGAN uses a multi-task discriminator network to perform two prediction tasks: an auxiliary network for predicting the thermal sensor label, and an adversarial network for predicting whether the generated thermal image is real or fake. We focus on the thermal sensor class prediction as an auxiliary task in an attempt to gain information about optical differences in images resulting from various thermal sensors such as sensitivity and resolution (e.g. uncooled microbolometer VOx sensor and BST ferroelectric sensor). favtGAN shows markedly better image quality results over pix2pix when combining data from face and cityscapes captured from the same thermal sensor. 
Our contributions are:
\begin{itemize}[leftmargin=*]
\item The first work to study VT translation of human faces, by developing a pix2pix-based favtGAN model.
    \item We study the image quality of generated thermal face images which is important for medical applications.
    \item We bootstrap training of image translation with additional data from different domains but similar thermal sensors to improve thermal image generation.
\end{itemize}
\begin{figure*}[t]
     \centering
     \begin{subfigure}[b]{0.24\textwidth}
         \centering
         \includegraphics[width=\textwidth]{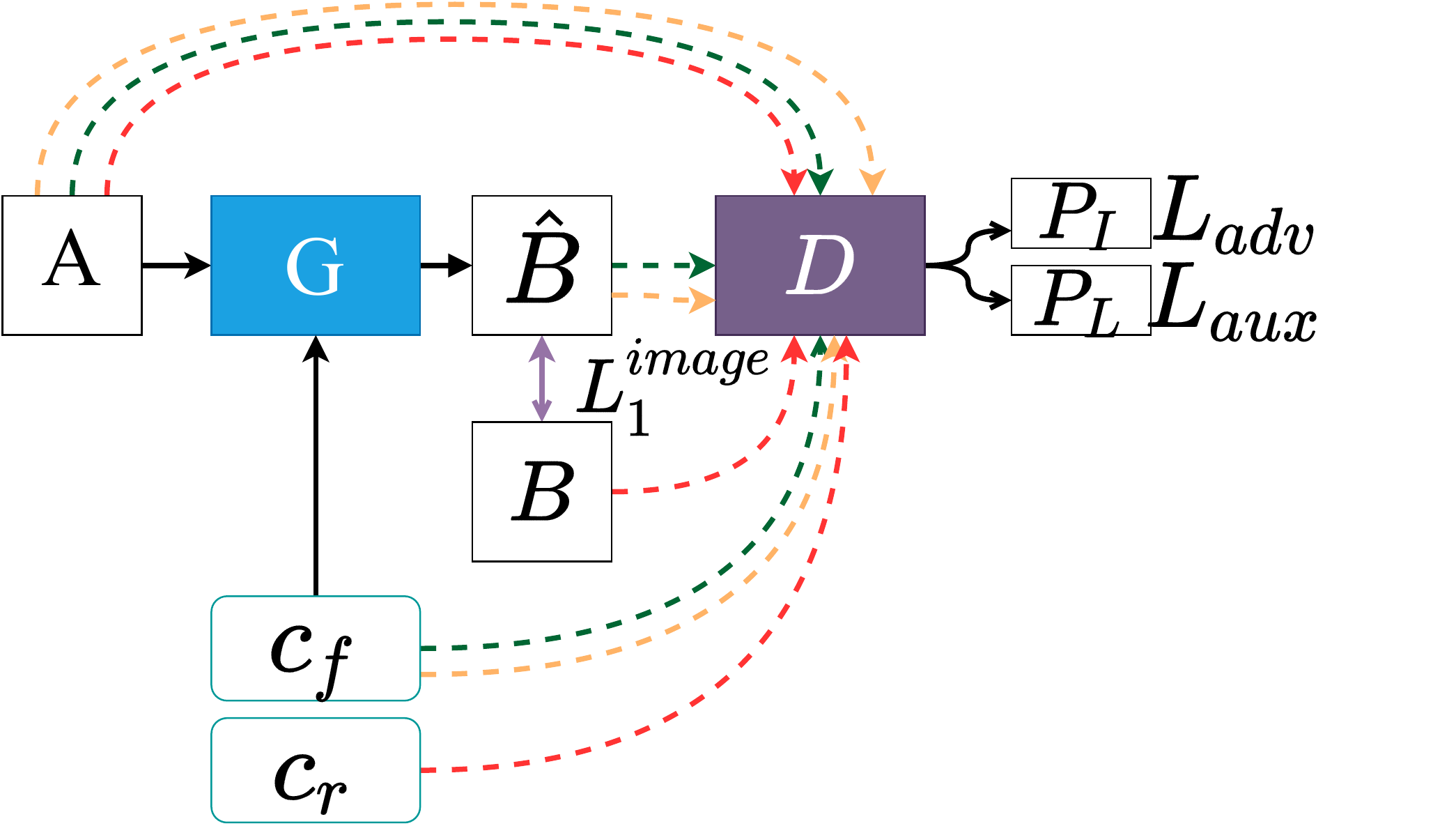}
         \caption{favtGAN}
         \label{baseline_arch}
     \end{subfigure}
    \begin{subfigure}[b]{0.24\textwidth}
        \centering
        \includegraphics[width=\textwidth]{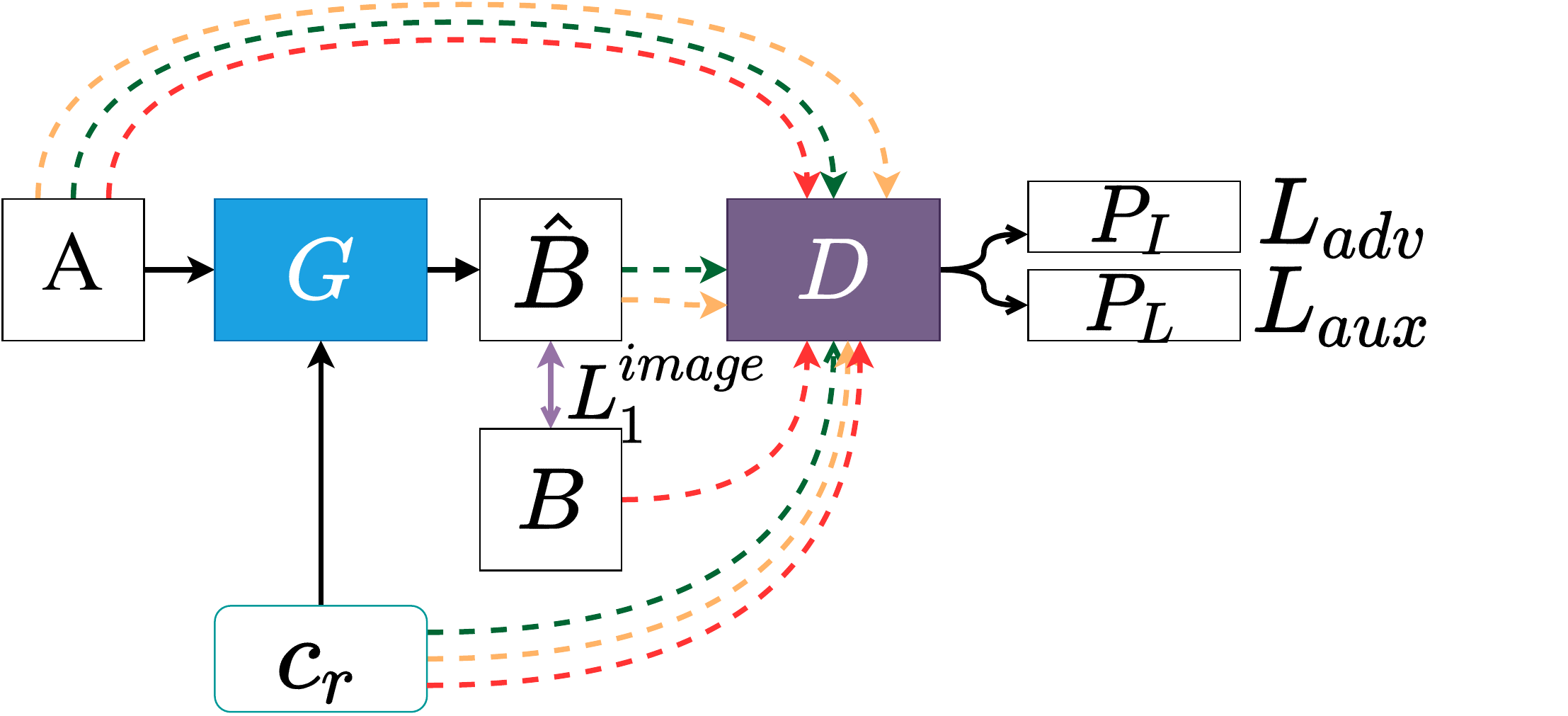}
        \caption{No Noise}
        \label{nonoise_arch}
     \end{subfigure}
    \begin{subfigure}[b]{0.24\textwidth}
        \centering
        \includegraphics[width=\textwidth]{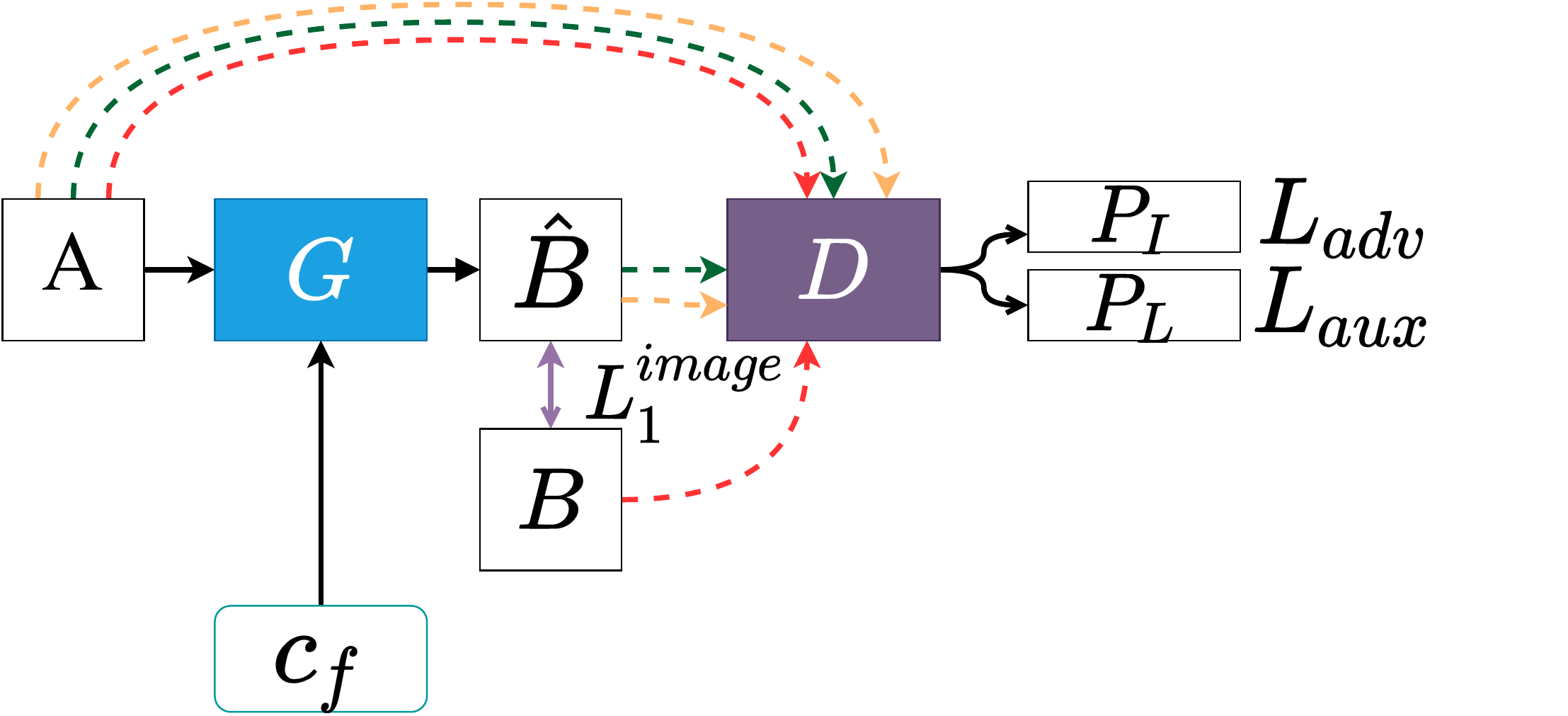}
        \caption{Noisy Labels}
        \label{noisy_labels}
     \end{subfigure}
    \begin{subfigure}[b]{0.24\textwidth}
        \centering
        \includegraphics[width=\textwidth]{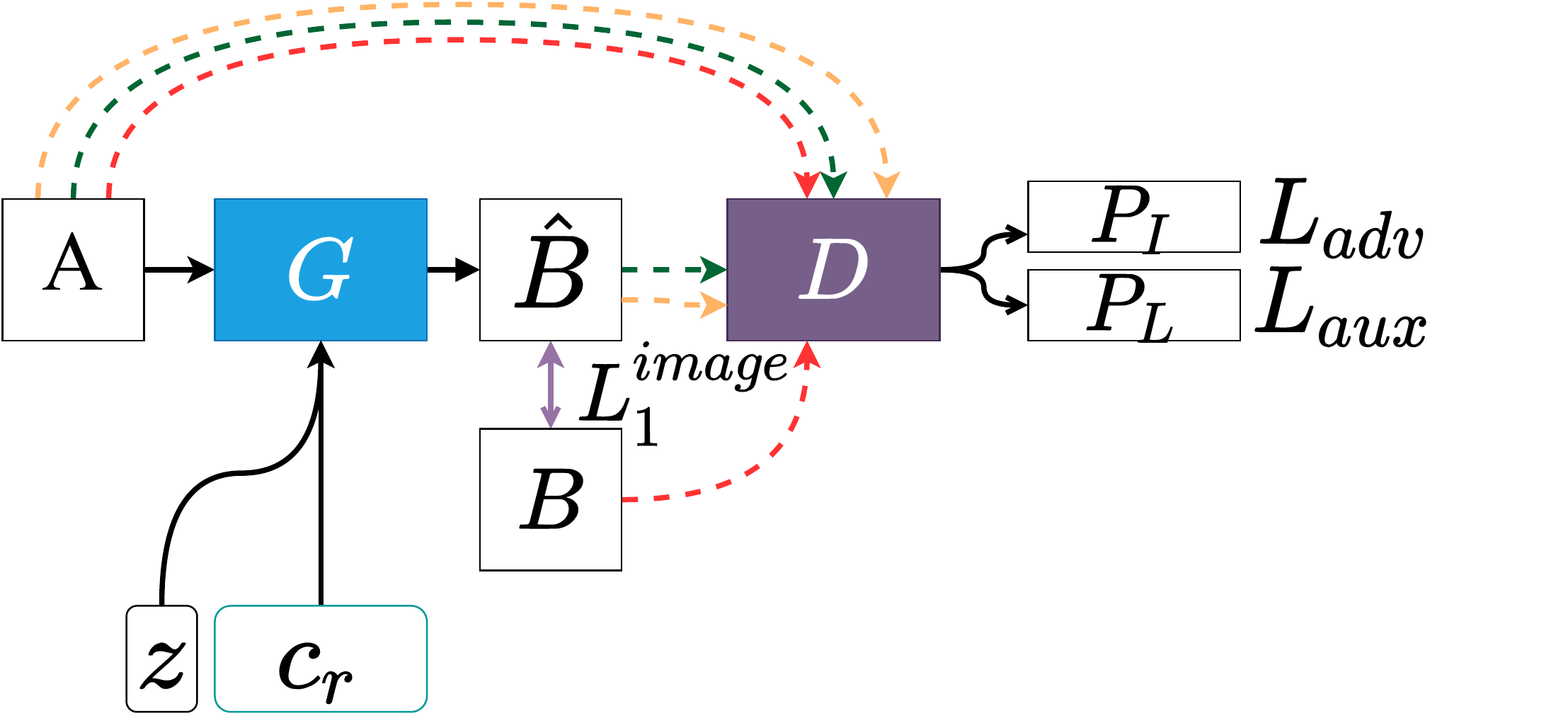}
        \caption{Gaussian Noise}
        \label{gaussian_and_true}
     \end{subfigure}
        \caption{
        \small{
        \textbf{favtGAN and Three Variations for Using the Thermal Side Labels as Noise and/or Real Inputs to the Discriminator}. (a) favtGAN Baseline model where $G$ is conditioned on ``noisy labels",$c_f$, in the form of discrete integers (0, 1, 2), and $D$ is conditioned on real true sensor labels, $c_r$ in the same set of integers; (b) ``No Noise" variation conditions $G$ and $D$ only on $c_r$; (c) ``Noisy Labels" variation conditions $G$, only, on $c_f$; (d) ```Gaussian" variation conditions $G$, only, on both Gaussian noise, $z$, and $c_r$. Green, red, and yellow lines indicate three adversarial losses and three auxiliary losses - one for the $G$ (green), real $D$ (red), and fake $D$ (yellow). $P_I$ refers to the predicted image (1, 16, 16) tensor (fake or real) and $P_L$ refers to the predicted sensor label. Full symbol notation is in the Approach.}}
        \label{arch}
\end{figure*}

\subsection{Related Work}
\label{related_work}
Several TV image-to-image translation GANs operate in the $thermal \rightarrow visible$ (TV) direction. These are not used for medical applications, but rather law enforcement and person re-identification focused on reconstructing the visible identity of persons from thermal surveillance images and thermal face recognition \cite{zhang2018tv, zhang2019synthesis, babu2020pcsgan, lai2019multi,wang2018thermal,chen2019matching, chu2018parametric}. Further, the TV GAN approaches we identified are trained on one single dataset collected by a single thermal sensor \cite{ioannou2014thermal,selinger2006appearance,buddharaju2007physiology,pavlidis2000imaging,wilder1996comparison, goulart2019emotion, puri2005stresscam, zhang2019synthesis, babu2020pcsgan}.  ThermalGAN \cite{kniaz2018thermalgan} is a pipeline where the first generator is BicycleGAN \cite{zhu2017toward} with some slight modifications to U-NET \cite{ronneberger2015u}, that outputs multi-modal images of segmented masks, followed by a second generator using pix2pix \cite{isola2017image}. ThermalGAN is trained and tested on a single dataset from one sensor, and is focused not on faces but the entire body for surveillance applications. ``TV-GAN" \cite{zhang2018tv} also uses pix2pix without modification, and implements a closed set recognition identity task using log loss. Side information has been used in TV GANs to explicitly regularize networks, where Zhang et al. \cite{zhang2018tv} uses one-hot encoding of subject identifiers, Kniaz et al. \cite{kniaz2018thermalgan} uses temperature vectors, and Chen and Ross \cite{chen2019matching} use facial labels (i.e left eye, right eye, right brow, etc.).   In another approach, Zhang et al \cite{zhang2019synthesis} explicitly feature-engineer their design for polarimetric facial TV translation.  Their approach involves preservation of texture and geometric features by leveraging multiple thermal polarimetric modalities, through feature-level fusion of Stokes images. Li et al. generates thermal pedestrian landscapes from visible images as a data augmentation tool for a downstream object detection task \cite{li2020unsupervised}. They apply CycleGAN \cite{zhu2017unpaired} for the landscape generation followed by an intensity inversion transformation \cite{li2020unsupervised}. The TV methods are different from our approach as they customize their networks with perceptual loss to improve the resolution quality, specific for data generated from one thermal sensor type. Their primary goals are to guarantee the identity from synthesized visible faces. Whereas, in our approach we seek to combine data from multiple thermal sensors in order to reconstruct thermal facial features. While the work by Li, et al. is similar, their focus is on the entire body and pedestrian landscapes versus close-up thermal face translation which makes our approach different.

\section{APPROACH}
Our approach, favtGAN, modifies the pix2pix image-to-image translation framework by conditioning the generator and discriminator on thermal sensor labels shown in Fig.1a. We refer to the favtGAN model as the ``favtGAN Baseline" for the rest of the paper. The discriminator is a multi-task network that outputs fake or real thermal patch probabilities used to calculate an adversarial loss, and probabilities to calculate an auxiliary loss that predicts the thermal sensor label. We design three variations of favtGAN shown in Fig.1b-d, to test the impact of using fake sensor labels, $c_f$, or $c_r$, real sensor labels to condition the network. Notation for Fig. 1 and formulas that follow are: $G$ is the generator, $D$ is the discriminator, $c_f$ are noisy labels which are random discrete integers in the set of thermal sensor labels (i.e. 0, 1, 2), $c_r$ are the real thermal sensor labels, $A$ is the real visible image, $B$ is the real thermal image, and $\hat{B}$ is the generated thermal image. Our code is available at {\small \url{https://github.com/umbc-sanjaylab/favtGAN}}.

\vspace{-10pt}
\subsection{Generator}
\vspace{-5pt}
We use U-NET \cite{ronneberger2015u} as a single generator for favtGAN. The generator uses both the noisy class label $c_f$ and visible image $A$ to generate images $\hat{B} = G(A, c_f)$.The total generator loss in Eq.4 is a composition of the adversarial loss in Eq.1., auxiliary (cross entropy) loss Eq.2, and the $L_{1}^{\mathit{image}}$ pixel reconstruction loss shown in Eq.3. Note that in Eq.1, we adopt the least-squares GAN approach (LSGAN) \cite{mao2017least} as the generator adversarial loss between the discriminator output of $D(A, \hat{B}, c_f)$ and 1, the real adversarial label. We applied one-sided label smoothing where 1 is replaced with 0.9 per ~\cite{chintala_2016}. 

\small

\setlength{\belowdisplayskip}{-1pt} \setlength{\belowdisplayshortskip}{-3pt}
\setlength{\abovedisplayskip}{-2pt} \setlength{\abovedisplayshortskip}{-2pt}


\begin{equation} \label{eq1}
\begin{split}
L_{\mathit{Adv}}(G) = \frac{1}{2}\mathop{\E}_{A\sim p_\mathit{vis}, c_f \sim U\{0,1\}, \hat{B} \sim p_G}[(D(A, \hat{B}, c_f) - 1)^2]
\end{split}
\end{equation}


\begin{equation} \label{eq2}
L_{\mathit{aux}}(G) = \mathop{\E}_{A\sim p_\mathit{vis}, c_f \sim U\{0,1\}, \hat{B} \sim p_G}[\log C(A, \hat{B}, c_f)]
\end{equation}


\begin{equation} \label{eq3}
L_1^{\mathit{image}}(G) = \\
\mathop{\E}_{B \sim p_\mathit{thr}, \hat{B} \sim p_G}  \| B - \hat{B})\|_1
\end{equation}

\begin{equation} \label{eq4}
\begin{split}
L_{G} = L_{\mathit{Adv}}(G) + L_{\mathit{aux}}(G) + \lambda L_1^{\mathit{image}}(G)
\end{split}
\end{equation}

\normalsize

\subsection{Discriminator}
\vspace{-5pt}
We use the same architecture as the pix2pix PatchGAN discriminator but modify it by adding a second layer that performs sensor class label prediction using softmax activation. As a result, the discriminator takes as input the visible image, thermal image, and the sensor label. Using this input, our discriminator conducts the adversarial task of  predicting real versus fake on a 16 x 16 patch across a 256 x 256 input  and second, a classification task to predict the sensor class labels $c$ for the input. The adversarial real and fake discriminator losses are shown in Eq.5 and Eq.6, respectively, using the LSGAN approach \cite{mao2017least}. Auxiliary losses are shown in Eqs. 7 and 8. The total discriminator loss is shown in Eq.9. The full objective function for favtGAN is given by Eq.10, and solved using the Adam optimizer.

\small

\setlength{\belowdisplayskip}{-1pt} \setlength{\belowdisplayshortskip}{-2pt}
\setlength{\abovedisplayskip}{-3pt} \setlength{\abovedisplayshortskip}{-3pt}


\begin{equation} \label{eq5}
L_{\mathit{Adv}_{\mathit{D_{real}}}} = \frac{1}{2}\mathop{\E}_{A \sim p_{vis},B \sim p_{thr},c_r \sim p_\mathit{thr}^{l}}\left[(D(A, B, c_r) - 1)^2\right]
\end{equation}


\begin{equation} \label{eq6}
L_{\mathit{Adv}_{\mathit{D_{fake}}}} = \frac{1}{2}\mathop{\E}_{A\sim p_\mathit{vis}, c_f \sim U\{0,1\}, \hat{B} \sim p_G}\left[(D(A, \hat{B},c_f) - 0)^2\right]  
\end{equation}


\begin{equation} \label{eq7}
L_{\mathit{aux}_{\mathit{D_{real}}}} = \mathop{\E}_{A \sim p_{vis}, B \sim p_{thr},c_r\sim p_\mathit{thr}^{l}}\left[\log C(A, B, c_r)\right] 
\end{equation}


\begin{equation} \label{eq8}
L_{\mathit{aux}_{\mathit{D_{fake}}}} = \mathop{\E}_{A\sim p_\mathit{vis}, c_f \sim U\{0,1\}, \hat{B} \sim p_G}\left[\log C(A, \hat{B},c_f)\right]
\end{equation}

\begin{equation} \label{eq9}
\begin{split}
L_{D} = \frac{1}{2}\big[(L_{\mathit{Adv}_{\mathit{D_{real}}}} + L_{\mathit{aux}_{\mathit{D_{real}}}}) 
+ (L_{\mathit{Adv}_{\mathit{D_{fake}}}} + L_{\mathit{aux}_{\mathit{D_{fake}}}})\big] 
\end{split}
\end{equation}

\begin{equation} \label{eq10}
G^* = \min_G L_{G} + \min_D  L_{D} 
\end{equation}

\normalsize

\begin{figure*}[t]
     \centering
     \begin{subfigure}[b]{0.48\textwidth}
         \centering
         \includegraphics[width=\textwidth]{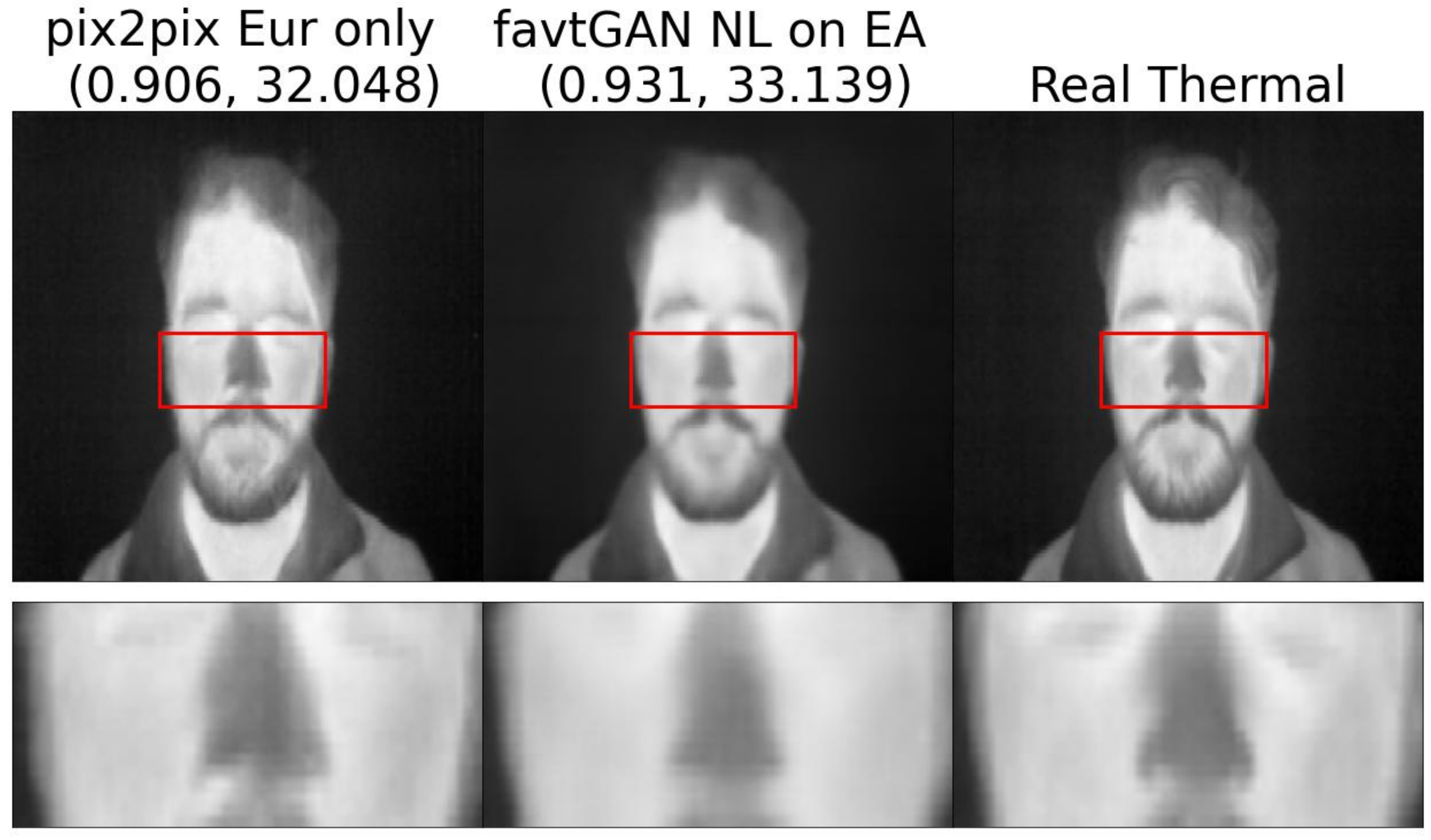}
     \end{subfigure}
    \begin{subfigure}[b]{0.48\textwidth}
        \centering
        \includegraphics[width=\textwidth]{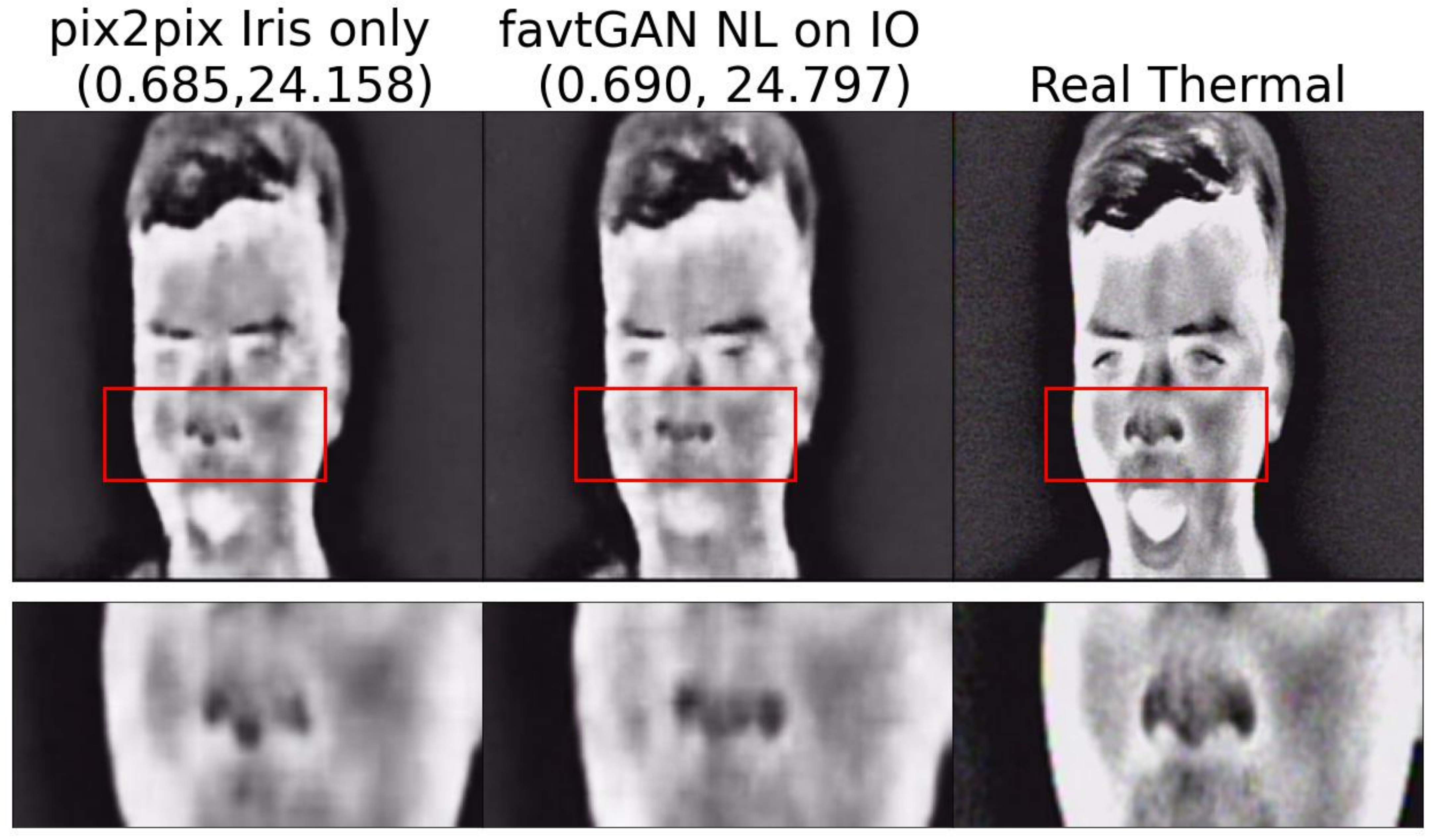}
     \end{subfigure}
     \small
     \caption{\small{\textbf{Generated Thermal Images Translated from the Visible Test Set.} Samples are shown from the best performing favtGAN experiment trained on combined face and cityscape datasets, compared to training pix2pix on a single dataset. Average SSIM and PSNR scores are provided. Red boxes show regions of interest, which are magnified in the second row. ``NL": Fig.1c Noisy Labels variation.}}
\end{figure*}

\section{EXPERIMENTS}
\vspace{-3pt}
\subsection{Datasets}
\vspace{-5pt}
\label{dataset_section}
We use four paired thermal-visible image datasets: Eurecom \cite{mallat2018benchmark}, FLIR ADAS \cite{flir_adas_readme}, Iris \cite{iris}, and Oklahoma State University (OSU) Color-Thermal \cite{Davis_Sharma}. From 2100 images, the Eurecom dataset was converted into a paired dataset of 1050 images (945 train images and 45 train subjects, 105 test images and 5 test subjects). The Iris dataset consists of faces with expressions (mouth open, smiling, mouth closed), and five illumination rotations. We only use expressions leading to 944 paired images (846 train images and 26 train subjects, 98 test images and 3 test subjects). We used the Advanced Driver Assistance Systems (ADAS) made by Flir \cite{flir_adas_readme} and randomly sampled images to make 940 pairs (842 train images, 98 test images) to balance against Iris and Eurecom. We also used the OSU dataset \cite{Davis_Sharma} consisting of 1054 pairs (843 train images, 211 test images) taken from ``Sequence 1". Eurecom and ADAS thermal images are captured by Flir cameras which use microbolometer sensors. Iris and OSU used a Raytheon Palm-IR-Pro and Raytheon 300D thermal camera, respectively, which are BST ferroelectric sensors.
\vspace{-3pt}
\label{sec:experimental_design}
\subsection{Experimental Design}
\vspace{-5pt}
There are no VT GAN translation architectures designed for thermal face translation, to our knowledge. As a result, we compared our approach to pix2pix, which has been used in both ``TV GAN" \cite{zhang2018tv} and ThermalGAN \cite{kniaz2018thermalgan}, in addition to three variations of the favtGAN architecture shown in Fig.2. Prior to selecting pix2pix as a comparative model, we also trained our datasets on CycleGAN \cite{zhu2017unpaired} and StarGAN \cite{choi2018stargan} which resulted in poor generated thermal images. We implement seven total experiments to generate Eurecom and Iris thermal faces. First, we train pix2pix without modification on only the single face dataset of either Eurecom or Iris to establish baseline SSIM and PSNR scores. We then train five experiments on the combined face datasets of Eurecom + Iris (EI) using favtGAN and pix2pix unmodified. The seventh experiment, indicated in bold in Table 1, trains the best performing favtGAN on different cityscape and human faces: ADAS + Eurecom (EA), and OSU + Iris (IO). We modified the implementation of the pix2pix GAN developed by Erik Lindernoren \cite{lindernoren_2018}. Models were trained using PyTorch on a RTX 8000 GPU, trained to 2000 epochs. Each experiment is set to a batch size of 12, Adam optimizer with a learning rate of 0.00002, b1 of 0.5, b2 of 0.999, decay of 100, 3 channels, and a lambda hyperparameter for the pixel reconstruction loss of 100. We use SSIM and PSNR to evaluate the image quality of the generated thermal faces since these metrics have been used to judge the quality of multispectral and near IR images from remote sensing, polarimetric face recognition, and biomedical applications \cite{zhang2019synthesis, kuang2020thermal, zhang2020feature, xu2020adversarial, perera2018in2i}. 

\section{RESULTS}
\vspace{-5pt}
\subsection{Quantitative Results}
\vspace{-5pt}
Results are shown in Table 1. We obtain the best results when training the favtGAN Noisy Labels variation on faces and cityscapes with similar thermal properties. The ADAS + Eurecom experiment trained on favtGAN Noisy Labels show 2.69\% increase of SSIM and 3.29\% increase of PSNR, over pix2pix on Eurecom alone. The same domain transfer effect occurs with Iris. Training the favtGAN Noisy Label implementation on OSU + Iris led to a 0.72\% increase of SSIM and 2.58\% increase of PSNR scores, over pix2pix trained on Iris, alone. But, we observe that training favtGAN Baseline on combined faces (Eurecom + Iris), even from different sensors, improves scores over training on Eurecom, alone, with a 2.09\% improvement in SSIM and 0.98\% improvement in PSNR. However, best results are seen when combining data from similar thermal sensors. Even if the domains are different, this may be a practical way to bootstrap training and transfer learning for VT face translation. These results are significant because improved image quality represents physiological accuracy of temperature patterns. The favtGAN variations that do not condition the generator on noise, but instead use the true sensor labels, $c_r$, show the lowest scores. 
\vspace{-10pt}
\subsection{Qualitative Results}
\label{qualitative}
\vspace{-5pt}
In Fig.2 we show samples comparing pix2pix trained on a single dataset and the best performing favtGAN architecture trained on the combined face and cityscape dataset. Pix2pix images trained on a single dataset alone, generate coarser textures and darker pixels than favtGAN. The red box shows the nose region which we observed had the greatest variability when qualitatively inspected. These are magnified in the bottom row. Pix2pix trained only on Eurecom shows structural deformities and smudging with less articulation than favtGAN. Iris images are more challenging where the differences are nuanced. The thermal image generated from pix2pix has slightly coarser texture on the right cheek compared to the favtGAN image. The nose regions are different where pix2pix generates lighter pixels and irregular structure than favtGAN. The placement of pixels, their contrast, hue, and distribution is important for thermal physiology because they are representative of facial temperatures. These temperatures are linked to different physiological responses and conditions \cite{ioannou2014thermal}. Therefore, failed reconstruction can lead to poor medical interpretation.


\begin{table}[t]
\caption{\small{\textbf{Image Quality Metrics using Mean SSIM and Mean PSNR for Generated Thermal Images, Translated from the Visible Test Set.} SSIM \% and PSNR \% show the relative change compared to pix2pix trained only a single face dataset. FG: favtGAN}}
\label{image_results}
\adjustbox{max width=\columnwidth}{%
\begin{tabular}{@{}llllll@{}}
\toprule
                        &                              & \multicolumn{4}{c}{\textbf{Eurecom}}                                   \\ \midrule
\textbf{Dataset}        & \textbf{Experiment}          & \textbf{SSIM}  & \textbf{PSNR}   & \textbf{SSIM \%} & \textbf{PSNR \%} \\
Eurecom                 & pix2pix                      & 0.906          & 32.048          & -                & -                \\
EI                      & pix2pix                      & 0.924          & 32.133          & 1.98\%           & 0.26\%           \\
EI                      & FG-Baseline             & 0.925          & 32.366          & 2.09\%           & 0.98\%           \\
EI                      & FG-No Noise             & 0.914          & 29.230          & 0.85\%           & -9.64\%          \\
EI                      & FG-Noisy Labels          & 0.925          & 31.835          & 2.02\%           & -0.67\%          \\
EI                      & FG-Gauss. Noise       & 0.909          & 28.242          & 0.36\%           & -13.48\%         \\
EA                      & FG-Baseline             & 0.931          & 33.104          & 2.69\%           & 3.19\%           \\
\textbf{EA}             & \textbf{FG-Noisy Labels} & \textbf{0.931} & \textbf{33.139} & \textbf{2.69\%}  & \textbf{3.29\%}  \\ \midrule
                        &                              & \multicolumn{4}{c}{\textbf{Iris}}                                      \\ \midrule
\textbf{Dataset}        & \textbf{Experiment}          & \textbf{SSIM}  & \textbf{PSNR}   & \textbf{SSIM \%} & \textbf{PSNR \%} \\ \midrule
Iris                    & pix2pix                      & 0.685          & 24.158          & -                & -                \\
EI                      & pix2pix                      & 0.681          & 23.946          & -0.54\%          & -0.89\%          \\
EI                      & FG-Baseline             & 0.682          & 24.060          & -0.37\%          & -0.41\%          \\
EI                      & FG-No Noise             & 0.653          & 22.000          & -4.91\%          & -9.81\%          \\
EI                      & FG-Noisy Labels          & 0.682          & 23.990          & -0.42\%          & -0.70\%          \\
EI                      & FG-Gauss. Noise       & 0.652          & 22.083          & -5.07\%          & -9.40\%          \\
IO                      & FG-Baseline             & 0.686          & 24.474          & 0.15\%           & 1.29\%           \\
\textbf{IO}             & \textbf{FG-Noisy Labels} & \textbf{0.690} & \textbf{24.797} & \textbf{0.72\%}  & \textbf{2.58\%}  \\ \bottomrule
\end{tabular}
}
\end{table}

\vspace{-15pt}
\section{CONCLUSION}
\label{sec:majhead}
\vspace{-5pt}
We have demonstrated the ability to implement visible-to-thermal image translation of human faces using favtGAN, a modified pix2pix network conditioned on thermal sensor class labels. Empirical results show that training favtGAN on combined faces and cityscapes data improves image quality if they share similar thermal sensor types. These results are preliminary and open up many new research questions which we plan to focus in our future work. In particular, can other domains from different thermal sensors improve favtGAN results?
We can advance the favtGAN architecture by explicitly applying domain adaptation of these varied domains, coupled with image generation. Finally, targeting the specific optical properties of long-wave infrared to improve translation between pixels and temperature is an important objective for future research.

\textbf{Acknowledgements:} 
This work is partially supported by grant IIS–1948399 from the US National Science Foundation and grant 80NSSC21M0027 from the National Aeronautics and Space Administration.
\vspace{-10pt}
\def\bibfont{\fontsize{9}{10}\selectfont}
\bibliographystyle{IEEEtranN}
\bibliography{references}

\end{document}